\documentclass{interact}
\usepackage{amsmath,amssymb}
\usepackage{booktabs}
\usepackage{array}
\usepackage{graphicx}
\usepackage{adjustbox}

\usepackage{tikz}
\usetikzlibrary{
  arrows.meta,
  positioning,
  shapes.geometric,
  fit,
  calc,
  backgrounds
}

\usepackage{pgfplots}
\pgfplotsset{compat=1.18}

\usepackage{longtable}
\usepackage{enumitem}
\usepackage[round,authoryear]{natbib}

\usepackage{xurl}
\usepackage[hidelinks]{hyperref}

\newcommand{\claire}{CLAIRE}
\newcommand{\best}[1]{\textbf{#1}}

\articletype{RESEARCH ARTICLE}
\title{CLAIRE: A Schema-Grounded Hybrid Workflow for Healthcare Administrative Form Completion}

\author{%
  \name{Garapati Keerthana\textsuperscript{1}\thanks{CONTACT Garapati Keerthana. Email: p20240505@hyderabad.bits-pilani.ac.in} and Manik Gupta\textsuperscript{1}}%
  \affil{\textsuperscript{1}Department of Computer Science and Information Systems, Birla Institute of Technology and Science, Pilani, Hyderabad, India}%
}
\received{}

\begin{document}

\maketitle

\begin{abstract}
Healthcare administrative staff transfer structured information from electronic health records, referrals, claims systems, provider rosters, and work queues into dynamic forms. We developed and evaluated \claire{} (Clinical Language and Agentic Intelligence for Reasoning and Entry), a hybrid workflow that separates field-state discovery, source-to-field mapping, deterministic validation, bounded correction, escalation, and audit tracing. We tested five synthetic healthcare administrative schemas, 1,000 source records, four interface variants, two data-quality suites, and six comparators, yielding 24,000 benchmark episodes. A separate strict-output audit evaluated direct mappings from Qwen2.5-1.5B and Qwen2.5-7B, and a trace-derived operational simulation covered 6,000 episodes. Under the evaluated synthetic benchmark conditions, full \claire{} achieved 1.000 episode success, field accuracy, required-field completion, and dependency completion in both suites; removing validation reduced stress-suite success to 0.500. In the simulation, 100.0\% of clean and validation-stress episodes reached a staff-reviewable draft, compared with 68.6\% of escalation-challenge episodes; unsupported cases were blocked. Scenario-based savings were 149.7--165.5 seconds per case, not observed staff times. The findings support schema-grounded, validation-first healthcare administrative automation in which language-model components assist mapping but do not authorize unsupported or consequential actions.
\end{abstract}


\section*{Highlights}
\begin{itemize}[leftmargin=*]
  \item Evaluated schema-grounded agentic form completion for healthcare operations.
  \item Schema grounding improves robustness across perturbed forms.
  \item Validation routes routine, review, and escalation cases.
  \item Simulation estimated 149.7--165.5 seconds saved per case.
\end{itemize}

\section{Introduction}

Healthcare staff repeatedly transfer information from electronic health records (EHRs), referrals, claims, provider rosters, and work queues into forms for intake, benefit verification, prior authorization, credentialing, and claims correction. Missing or inconsistent entries can delay care and create rework, making administrative form completion a medical-informatics problem \cite{shrank2019waste}. The informatics challenge is to preserve field meaning, dependencies, validation constraints, provenance, and human oversight as information moves from heterogeneous healthcare sources into structured interfaces. Current US policy similarly promotes computable exchange and reduced prior-authorization burden \cite{cms2024priorauth}.

These forms are stateful: one answer can reveal another required field, which we call a \emph{dependency}. Automation must find visible fields, reveal dependencies, map only source-supported values, validate the completed state, and stop when evidence is missing. For example, it may normalize a supported National Provider Identifier (NPI) to the required ten-digit format \cite{cms2024npi}, but must not invent one. Fixed browser scripts can be brittle when identifiers, labels, or layouts change, whereas unconstrained language models can produce unsupported, malformed, or untraceable entries \citep{selenium2026,playwright2026,puppeteer2026}.

Web-agent benchmarks show that flexible interpretation has not yet produced dependable end-to-end action \citep{deng2023mind2web,zhou2023webarena,koh2024visualwebarena,he2024webvoyager,xie2024osworld}. FormFactory likewise identifies field--value matching, layout understanding, and precise interface action as persistent difficulties \citep{li2025formfactory}. Healthcare administration therefore needs a middle path: flexible mapping within explicit schemas, state tracking, validation, bounded correction, audit logging, and staff review. A \emph{form schema} is a machine-readable description of fields, labels, allowed values, dependencies, and rules; an \emph{episode} is one complete form attempt under one test condition.

We introduce \claire{} (Clinical Language and Agentic Intelligence for Reasoning and Entry), a schema-grounded hybrid architecture with four logical roles. The \textbf{Selector} finds visible fields and dependencies; the \textbf{Filler} maps supported values; the \textbf{Validator} checks the resulting state; and the \textbf{Error Corrector} applies only deterministic normalization or an explicitly supplied correction. Unresolved cases are blocked for staff review, and every step is logged. These are LLM-instrumented hybrid modules, not four autonomous LLM agents: deterministic schema and rule logic authorized actions in the main benchmark, while direct model-generated mappings were evaluated separately.

This study makes four contributions:
\begin{enumerate}[leftmargin=*,label=(\arabic*)]
  \item a four-role, schema-grounded healthcare informatics architecture with explicit audit and staff-review boundaries;
  \item a reproducible synthetic dataset spanning five form families, two data-quality suites, four interface variants, dependencies, and validation rules;
  \item six controlled comparators and a two-model strict-output audit separating model mapping from deterministic safeguards; and
  \item a trace-derived simulation of review readiness, escalation, and scenario-based staff time.
\end{enumerate}

\begin{table}[!htbp]
\centering
\caption{Summary table.}
\label{tab:summary}
\begin{tabular}{>{\raggedright\arraybackslash}p{0.28\linewidth}>{\raggedright\arraybackslash}p{0.62\linewidth}}
\toprule
What was already known & Healthcare organizations rely on administrative forms for prior authorization, benefit verification, credentialing, intake, and claims work. Browser automation can be brittle, and unconstrained language-model agents can generate plausible but unchecked entries. \\
\midrule
What this study added & A schema-grounded hybrid workflow separates field discovery, mapping, validation, controlled error correction, and audit logging. In 24,000 benchmark episodes and a 6,000-episode operational simulation, the workflow completed validated cases, intercepted review needs, captured escalation cases, and produced scenario-based estimates of staff-time savings. \\
\bottomrule
\end{tabular}
\end{table}

\clearpage

\section{Related work}
Interactive healthcare form completion sits between web agents, document AI, and administrative AI because it requires semantic mapping as well as safe control of a changing form state.
\subsection{Web agents and computer-use benchmarks}

Mind2Web showed that raw webpage content often needs filtering before language-model use, while WebArena and VisualWebArena introduced reproducible functional and visually grounded tasks \citep{deng2023mind2web,zhou2023webarena,koh2024visualwebarena}. WebVoyager, WorkArena, BrowserGym, and OSWorld extend evaluation to real websites, enterprise workflows, standardized infrastructure, and desktop interaction \citep{he2024webvoyager,drouin2024workarena,lesellier2024browsergym,xie2024osworld}. Together they show that browser agents remain unreliable on realistic tasks. Evaluation guidance therefore recommends reproducibility and downstream usefulness, while safety benchmarks identify poor risk awareness and brittle robustness \citep{kapoor2024agents,zhang2024agentsafety}. We respond by reporting validation and execution behavior, not accuracy alone.

\subsection{Interactive form filling}

FormFactory is the closest benchmark: it tests field--value alignment and graphical actions across layouts and input types \citep{li2025formfactory}. Its results demonstrate that form filling requires more than copying: a system must align evidence with fields, choose appropriate controls, and leave the complete form valid. Unlike its zero-shot action setting, we evaluate a healthcare-specific modular workflow with computable schemas, deterministic validation, correction, and traces. FormFactory motivates our Labels-only comparator and supplies healthcare-like answer keys for a secondary alignment check. Our adapter tests semantic mappings rather than screenshot-coordinate trajectories, so we do not claim a full reproduction.

\subsection{Document AI and form prediction}

LayoutLM and LayoutLMv3 combine text and spatial layout for static document understanding, scanned forms, and visual question answering \citep{xu2020layoutlm,huang2022layoutlmv3}; LAFF predicts categorical values from historical submissions \citep{belgacem2023laff}. These complementary methods support extraction or suggestion but do not themselves reveal conditional controls, validate a dynamic form state, or block unsafe submission.

\subsection{Healthcare administrative AI}

AI-generated prior-authorization letters can retain clinical content yet omit administrative details, and HealthAdminBench reports low end-to-end reliability despite stronger subtasks \citep{awan2026priorauthletters,bedi2026healthadminbench}. Prior work therefore supports individual capabilities but leaves a healthcare-specific gap: dynamic form filling in which every value is source-grounded, rule-checked, corrected only within stated limits, and escalated when unresolved.

\section{Methods}

\subsection{Architecture and workflow framing}

The architecture receives a \emph{structured source record} containing the available case values and a \emph{computable form schema} defining fields, labels, aliases, source keys, input types, allowed options, required flags, dependencies, and validation rules. All roles use this shared evidence and form state.

\begin{figure}[!htbp]
\centering
\includegraphics[width=\textwidth]{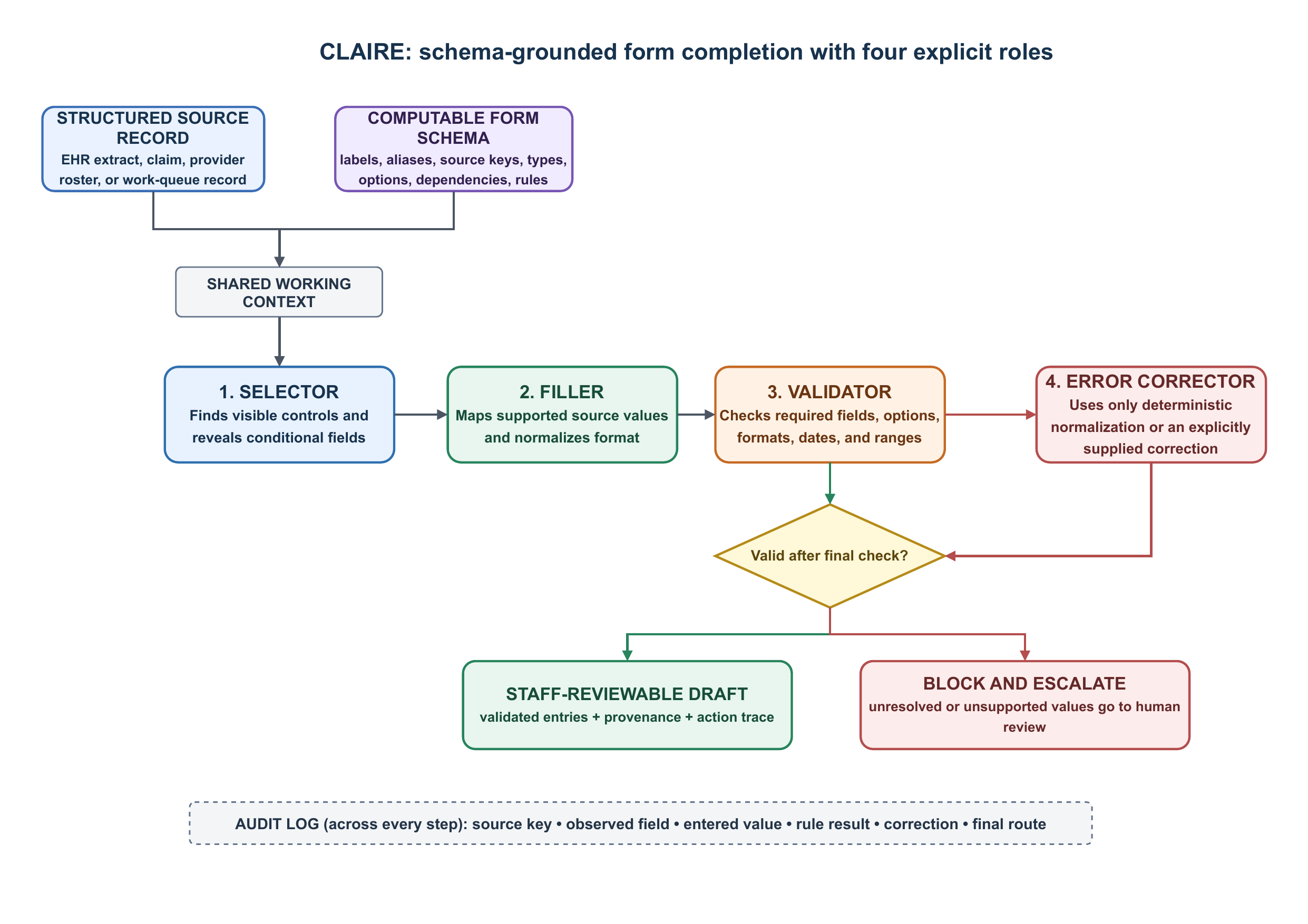}
\caption{Schema-grounded architecture. The four numbered roles share the same source record, schema, and working context. The Error Corrector uses only supported corrections, after which the Validator checks the state again. Unresolved cases are blocked for staff review.}
\label{fig:architecture}
\end{figure}

Four logical roles are implemented across SelectorBot, FillerBot, and ValidatorBot; error correction is a distinct ValidatorBot operation. The Selector finds visible controls and triggers known dependencies. The Filler maps source-supported values and normalizes formats. The Validator checks required fields, options, patterns, date order, ranges, and conditional rules. The Error Corrector may use deterministic normalization or an explicit corrected value already in the record, after which validation repeats. Valid cases become reviewable drafts; unresolved cases are blocked, and observations, actions, rule results, corrections, and routes are logged.

During the 24,000-episode benchmark, these modules issued prompts through Qwen/Qwen2.5-0.5B-Instruct and logged call metadata. The main executor used the model interface for role-specific prompting, but model-generated content did not authorize benchmark actions; execution was governed by schemas, deterministic matching, validation, and bounded correction. The environment and schema drove selection, deterministic matching drove filling, explicit rules drove validation, and evidence-bounded logic drove correction. Thus, the roles are LLM-instrumented hybrid agent modules rather than autonomous LLM agents. The strict-output audit separately used model-generated mappings as executable inputs. In intended use, \claire{} returns a draft and trace for staff review; it may reformat supported data but cannot invent facts, change a service code, or submit consequential forms independently (Figs.~\ref{fig:architecture} and \ref{fig:workflow}).

\subsection{Dataset and terminology}
Deterministic templates generated five form families---patient intake, prior authorization, benefit verification, provider credentialing, and claims resubmission---with one schema and 100 indexed records per family. Fixed rules varied synthetic names, identifiers, and dependency-triggering choices, so regeneration produces the same 500 clean records. No patient data, protected health information, or free-text clinical notes were used; the values exercise form logic rather than represent a clinical population.

Together, the 500 clean records and a 500-record validation-stress copy yielded 1,000 records. The stress copy inserted invalid NPIs and reversed dates into prior-authorization records, malformed claim identifiers and negative amounts into claims records, and invalid primary-care-provider NPIs into benefit records. Explicit corrected values were stored separately so correction could use supplied evidence instead of guessing; intake and credentialing remained unchanged controls. A \textbf{record} is one case, a \textbf{schema} defines one form family, an \textbf{interface variant} changes presentation without changing meaning, a \textbf{comparator} is an evaluated configuration, a \textbf{suite} groups a data condition, and an \textbf{episode} is one record--schema--variant--comparator run (Fig.~\ref{fig:benchmark-data}).

\subsection{Internal benchmark design and comparators}
The clean and validation-stress suites each contained 500 records, following synthetic-data and de-identification principles \cite{walonoski2018synthea,hhs2024deidentification}. Each record was tested with original labels, synonym labels, obfuscated identifiers, and shuffled layout to expose dependence on exact identifiers, wording, or order.

\begin{table}[htbp]
\centering
\caption{Internal benchmark design.}
\label{tab:design}
\begin{tabular}{>{\raggedright\arraybackslash}p{0.28\linewidth}>{\raggedright\arraybackslash}p{0.62\linewidth}}
\toprule
Component & Description \\
\midrule
Schemas & Patient intake, prior authorization, benefit verification, provider credentialing, and claims resubmission. \\
Records & 1,000 deterministic synthetic source records: 500 clean records and a 500-record validation-stress copy. \\
Interface variants & Original labels, synonym labels, obfuscated identifiers, and shuffled layout. \\
Comparators & Exact ID, Labels only, Monolithic, \claire{} without schema, \claire{} without the Validator, and full \claire{}. \\
Metrics & Episode success, field accuracy, required-field completion, dependency completion, validation errors, corrections, and model calls. \\
\bottomrule
\end{tabular}
\end{table}

The six comparators were Exact ID (direct identifier matches), Labels only (visible-label matching), Monolithic (one schema-aware step), \claire{} without schema, \claire{} without the Validator, and full \claire{}. Crossing 1,000 records, four variants, and six comparators produced 24,000 episodes (Fig.~\ref{fig:benchmark-data}; Table~\ref{tab:design}).

Episode success required every expected visible and conditionally revealed field to match the answer key with no final validation error. Field accuracy measured correct final values over expected fields; required-field and dependency completion measured whether mandatory and revealed controls were complete. We also recorded remaining validation errors, corrections, actions, and model calls. These complementary measures distinguish a form that contains several correct values from one that is safe and complete at the episode level.

\subsection{Operational simulation}
We ran a mathematical operational simulation over full \claire{} traces, not a usability study or live deployment. Clean and stress records under four variants supplied 2,000 episodes each. A third 2,000-episode escalation suite removed explicit corrections from fixed stress cases and the clinical reason from a fixed prior-authorization subset, testing whether unsupported cases were blocked. Specifically, corrected NPIs and dates were removed from selected prior-authorization records, corrected identifiers and amounts from claims records, and corrected NPIs from benefit-verification records. The construction tests the route taken when normalization is insufficient and no supported replacement remains.

For each trace, we counted fields, dependencies, Validator interceptions, corrections, and unresolved errors. Manual entry assumed 45 seconds to open a case, 18 seconds per field, 20 seconds per dependency, and 45 seconds per rework event. Assisted review assumed 20 seconds to open the trace, 4 seconds per field, 20 seconds per correction, 25 seconds after an interception, and 60 seconds plus 20 seconds per unresolved error for escalation.

These pre-specified round numbers make the scenario reproducible; they were not participant measurements or literature-derived per-action standards. Published evidence provides scale rather than validation: CAQH reports average prior-authorization transaction times of 24 minutes manually and 10 minutes electronically \cite{caqh2024index}, while a survey of 1,010 provider employees reported role-specific median burdens of 1--11 hours per week \cite{sahni2024priorauth}. The outputs are therefore scenario estimates, not observed productivity, and should be replaced with prospective time measurements and sensitivity analyses.

\subsection{Strict-output LLM audit and external alignment}
A strict-output audit isolated direct LLM mapping: Qwen2.5-1.5B and Qwen2.5-7B had to return parseable JSON connecting source values to fields without prose or Markdown \cite{yang2024qwen25}. Malformed output, unsupported mappings, incomplete dependencies, or final validation errors could therefore prevent episode success even when individual fields were correct. For each model and suite, 100 episodes were balanced across variants. Labels only supplied visible labels; Schema loop added schema context and repeated rounds; Schema + correction also exposed explicit corrections. Full \claire{} was the guarded reference. We measured success, field accuracy, parse validity, validation errors, and mapping agreement. A secondary check used 230 healthcare-like FormFactory answer-key examples; because it tested mappings rather than screenshot coordinates or action trajectories, Appendix~B reports alignment rather than full benchmark reproduction.

\section{Results}

\subsection{Internal benchmark comparison}
\begin{figure}[p]
\centering
\begin{tikzpicture}
\begin{axis}[
  width=\linewidth,
  height=8.4cm,
  ybar,
  bar width=7pt,
  ymin=0, ymax=1.08,
  ylabel={Episode success},
  symbolic x coords={Exact ID,Labels only,Monolithic,No schema,No validator,CLAIRE},
  xtick=data,
  x tick label style={rotate=28, anchor=east, font=\small},
  y tick label style={font=\small},
  label style={font=\large},
  ymajorgrids=true,
  grid style={draw=black!10},
  legend style={at={(0.5,1.04)}, anchor=south, legend columns=2, draw=none, font=\small},
  nodes near coords,
  nodes near coords style={font=\small, rotate=90, anchor=west},
  every axis plot/.append style={fill opacity=.88}
]
\addplot+[draw=blue!70!black, fill=blue!55] coordinates {
  (Exact ID,0.389) (Labels only,0.150) (Monolithic,0.518)
  (No schema,0.200) (No validator,1.000) (CLAIRE,1.000)
};
\addplot+[draw=orange!80!black, fill=orange!70] coordinates {
  (Exact ID,0.263) (Labels only,0.150) (Monolithic,0.350)
  (No schema,0.200) (No validator,0.500) (CLAIRE,1.000)
};
\legend{Clean dynamic forms, Validation stress}
\end{axis}
\end{tikzpicture}
\caption{Episode success in the full internal benchmark. \claire{} maintains complete success across clean and validation-stress suites, while validation removal specifically degrades performance under dirty source values.}
\label{fig:main-results}
\end{figure}
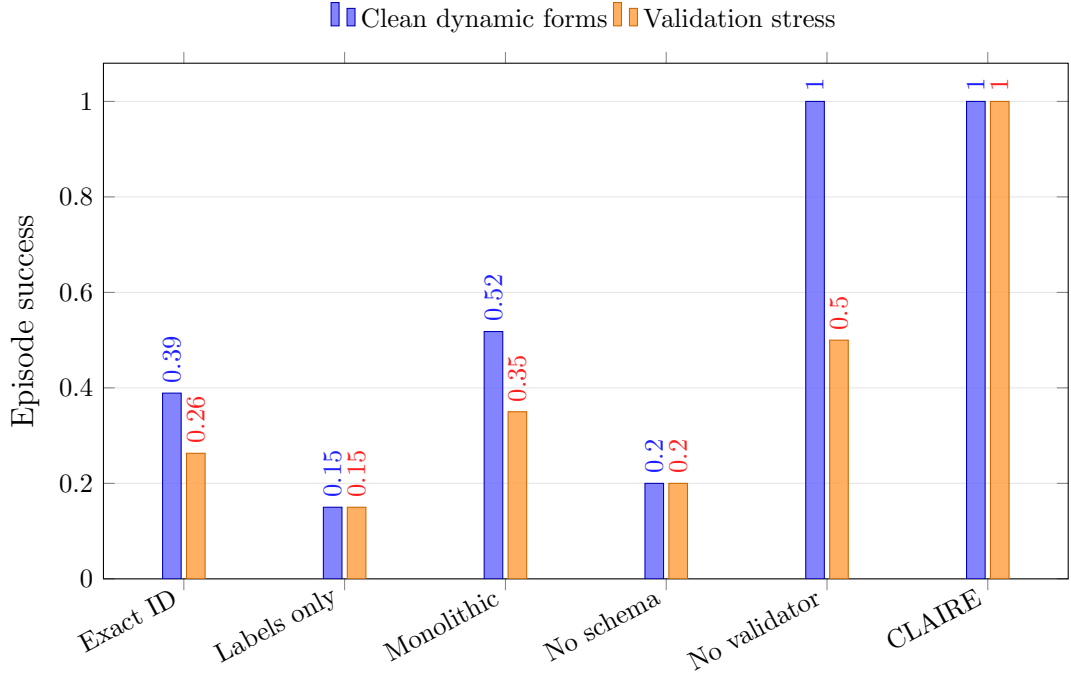

Across 24,000 internal episodes, full \claire{} achieved 1.000 episode success, field accuracy, required-field completion, and dependency completion in both suites, with zero final validation errors (Table~\ref{tab:results}; Fig.~\ref{fig:main-results}). These values mean that every expected visible and conditionally revealed field matched the answer key at the end of each controlled episode. They do not mean that the system is ready for arbitrary live portals; the result applies to the synthetic schemas, rules, and interface changes defined in Section~3.3.

\begin{table}[!htbp]
\centering
\caption{Internal benchmark results by suite. Lower values are better for validation errors.}
\label{tab:results}
\begin{adjustbox}{max width=\linewidth}
\begin{tabular}{llrrrrr}
\toprule
Suite & Comparator & Success & Field accuracy & Required & Dependency & Validation errors \\
\midrule
Clean & Exact ID & 0.389 & 0.705 & 0.700 & 0.518 & 1.912 \\
Clean & Labels only & 0.150 & 0.784 & 0.878 & 0.518 & 0.824 \\
Clean & Monolithic & 0.518 & 0.941 & 0.934 & 0.518 & 0.482 \\
Clean & \claire{} no schema & 0.200 & 0.857 & 0.960 & 0.942 & 0.250 \\
Clean & \claire{} no validator & \best{1.000} & \best{1.000} & \best{1.000} & \best{1.000} & \best{0.000} \\
Clean & \claire{} & \best{1.000} & \best{1.000} & \best{1.000} & \best{1.000} & \best{0.000} \\
\midrule
Stress & Exact ID & 0.263 & 0.634 & 0.700 & 0.518 & 2.512 \\
Stress & Labels only & 0.150 & 0.695 & 0.878 & 0.518 & 1.574 \\
Stress & Monolithic & 0.350 & 0.845 & 0.934 & 0.518 & 1.282 \\
Stress & \claire{} no schema & 0.200 & 0.863 & 0.966 & 0.942 & 0.200 \\
Stress & \claire{} no validator & 0.500 & 0.894 & \best{1.000} & \best{1.000} & 0.900 \\
Stress & \claire{} & \best{1.000} & \best{1.000} & \best{1.000} & \best{1.000} & \best{0.000} \\
\bottomrule
\end{tabular}
\end{adjustbox}
\end{table}

The comparator pattern explains the perfect full \claire{} score. Exact ID was disrupted by obfuscated identifiers. Labels only often filled individual fields correctly but left dependencies or validation errors unresolved, so its episode success was 0.150 in both suites. Monolithic schema-aware filling improved field accuracy but still missed dynamic completion. Removing schema hints reduced episode success to 0.200 in both suites. Removing the Validator preserved 1.000 clean-suite success but reduced validation-stress success to 0.500 and left 0.900 errors per episode. Thus, schema grounding helps map the right evidence, while validation and correction are specifically important when source values violate rules.

\subsection{Operational simulation}

The operational simulation produced three routes (Table~\ref{tab:operational}). In the 2,000 clean episodes, all cases reached a staff-reviewable draft; 48.2\% still triggered focused review because a Validator interception occurred in the trace. In the 2,000 validation-stress episodes, 65.0\% were intercepted and corrected before the final check, all reached a reviewable draft, and no final validation errors remained. In the 2,000 escalation-challenge episodes, 31.4\% were blocked, 68.6\% reached a draft, unsafe-case capture was 100.0\%, and the false-reassurance rate was 0.0\%. In plain terms, correctable cases were fixed using supplied evidence, while cases without that evidence were held for staff.

Under the pre-specified timing scenario, the estimated savings were 149.7 seconds per clean case, 165.5 seconds per validation-stress case, and 152.2 seconds per escalation-challenge case. These are calculated scenario outputs, not observed staff times. The more defensible operational finding is the routing behavior---routine draft, focused review, or escalation---because it follows directly from trace and validation state; the time estimates require prospective measurement.

\begin{table}[!htbp]
\centering
\large
\caption{Operational simulation metrics from 6,000 healthcare administrative form episodes. Time values are calculated from trace-derived field counts, validation events, correction events, and explicit time-motion assumptions.}
\label{tab:operational}
\begin{adjustbox}{max width=\linewidth}
\begin{tabular}{lrrrrrrr}
\toprule
Suite & Episodes & Ready & Focused review & Escalated & Intercepted & Time saved/case & Hours saved \\
\midrule
Clean dynamic forms & 2000 & 1.000 & 0.482 & 0.000 & 0.482 & 149.7s (68.6\%) & 83.2 \\
Validation stress & 2000 & 1.000 & 0.650 & 0.000 & 0.650 & 165.5s (65.8\%) & 92.0 \\
Escalation challenge & 2000 & 0.686 & 0.336 & 0.314 & 0.650 & 152.2s (61.6\%) & 84.6 \\
\bottomrule
\end{tabular}
\end{adjustbox}
\end{table}

\subsection{Strict-output LLM audit}
The strict-output audit asks a narrower question than the internal benchmark: can an LLM return the correct field--value mapping in machine-readable JSON when it receives labels alone, schema context, or schema context plus explicit correction? The result is counted at the episode level only when the final mapped state is complete and valid.

\begin{figure}[p]
\centering
\begin{tikzpicture}
\begin{axis}[
  width=\linewidth,
  height=7.25cm,
  ybar,
  bar width=7pt,
  ymin=0, ymax=1.08,
  ylabel={Episode success},
  symbolic x coords={Labels,Schema,Schema+correction,Guarded CLAIRE},
  xtick=data,
  x tick label style={rotate=20, anchor=east, font=\small},
  y tick label style={font=\small},
  label style={font=\large},
  ymajorgrids=true,
  grid style={draw=black!10},
  legend style={at={(0.5,1.04)}, anchor=south, legend columns=2, draw=none, font=\small},
  nodes near coords,
  nodes near coords style={font=\small, rotate=90, anchor=west},
  every axis plot/.append style={fill opacity=.88}
]
\addplot+[draw=teal!70!black, fill=teal!55] coordinates {
  (Labels,0.310) (Schema,0.810) (Schema+correction,0.810) (Guarded CLAIRE,1.000)
};
\addplot+[draw=purple!70!black, fill=purple!55] coordinates {
  (Labels,0.230) (Schema,0.480) (Schema+correction,0.900) (Guarded CLAIRE,1.000)
};
\legend{Qwen2.5-1.5B clean, Qwen2.5-1.5B stress}
\end{axis}
\end{tikzpicture}

\vspace{0.35cm}

\begin{tikzpicture}
\begin{axis}[
  width=\linewidth,
  height=7.25cm,
  ybar,
  bar width=7pt,
  ymin=0, ymax=1.08,
  ylabel={Episode success},
  symbolic x coords={Labels,Schema,Schema+correction,Guarded CLAIRE},
  xtick=data,
  x tick label style={rotate=20, anchor=east, font=\small},
  y tick label style={font=\small},
  label style={font=\large},
  ymajorgrids=true,
  grid style={draw=black!10},
  legend style={at={(0.5,1.04)}, anchor=south, legend columns=2, draw=none, font=\small},
  nodes near coords,
  nodes near coords style={font=\small, rotate=90, anchor=west},
  every axis plot/.append style={fill opacity=.88}
]
\addplot+[draw=blue!70!black, fill=blue!52] coordinates {
  (Labels,0.260) (Schema,0.650) (Schema+correction,0.650) (Guarded CLAIRE,1.000)
};
\addplot+[draw=red!70!black, fill=red!55] coordinates {
  (Labels,0.290) (Schema,0.840) (Schema+correction,0.840) (Guarded CLAIRE,1.000)
};
\legend{Qwen2.5-7B clean, Qwen2.5-7B stress}
\end{axis}
\end{tikzpicture}
\caption{Strict LLM decision audit on a focused subset. Direct labels-only mappings are weak; schema conditioning and explicit correction improve performance substantially, but the guarded CLAIRE executor remains the most reliable condition.}
\label{fig:strict-audit}
\end{figure}
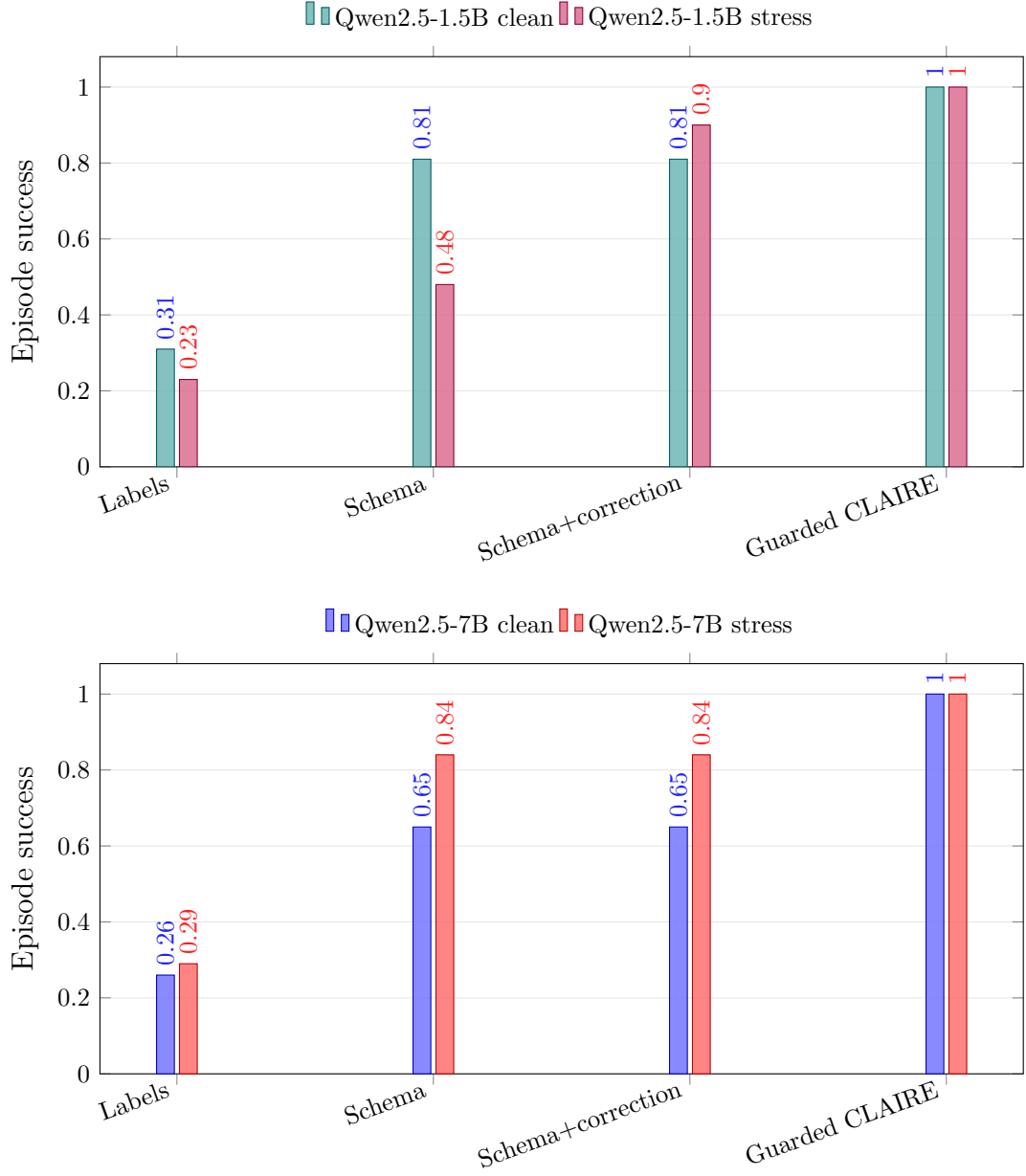

Both models benefited from structure, but neither direct-model condition matched the guarded executor across both suites (Table~\ref{tab:audit}; Fig.~\ref{fig:strict-audit}). For Qwen2.5-1.5B, validation-stress success rose from 0.230 with labels only to 0.480 with the schema loop and 0.900 with schema plus correction. For Qwen2.5-7B, schema-conditioned stress success was 0.840, with no additional success gain from explicit correction in this sample. Full \claire{} reached 1.000 because mapping was followed by deterministic checks and controlled correction. The practical conclusion is not that an LLM is unnecessary; it is that flexible mapping should not be treated as the final safety decision.

\begin{table}[!htbp]
\centering
\small
\caption{Focused strict LLM audit. Values are averaged across four variants and 100 episodes per suite, model, and condition.}
\label{tab:audit}
\begin{adjustbox}{max width=\linewidth}
\begin{tabular}{lllrr}
\toprule
Model & Suite & Condition & Success & Field accuracy \\
\midrule
Qwen2.5-1.5B & Clean & Labels only & 0.310 & 0.883 \\
Qwen2.5-1.5B & Clean & Schema loop & 0.810 & 0.926 \\
Qwen2.5-1.5B & Clean & Schema + correction & 0.810 & 0.926 \\
Qwen2.5-1.5B & Stress & Labels only & 0.230 & 0.779 \\
Qwen2.5-1.5B & Stress & Schema loop & 0.480 & 0.864 \\
Qwen2.5-1.5B & Stress & Schema + correction & 0.900 & 0.967 \\
Qwen2.5-7B & Clean & Labels only & 0.260 & 0.830 \\
Qwen2.5-7B & Clean & Schema loop & 0.650 & 0.952 \\
Qwen2.5-7B & Clean & Schema + correction & 0.650 & 0.952 \\
Qwen2.5-7B & Stress & Labels only & 0.290 & 0.881 \\
Qwen2.5-7B & Stress & Schema loop & 0.840 & 0.975 \\
Qwen2.5-7B & Stress & Schema + correction & 0.840 & 0.976 \\
Guarded executor & Both & Full \claire{} & \best{1.000} & \best{1.000} \\
\bottomrule
\end{tabular}
\end{adjustbox}
\end{table}

\section{Discussion}

The results support constrained autonomy with workflow accountability. Removing schema context reduced success to 0.200; removing validation reduced stress-suite success from 1.000 to 0.500 while leaving clean performance unchanged; and direct LLM mapping improved with schema context but remained less dependable than guarded execution. Discovery, validation, evidence-bounded correction, escalation, and traces therefore contributed distinct safeguards rather than acting as interchangeable modules. This differs from benchmarks that score only plausible output and from robotic process automation that assumes stable selectors.

The operational simulation extended correctness into workflow routing by separating routine drafts, focused review, and blocked escalation. This distinction resembles administrative work queues, where supported cases can proceed and exceptions require attention. However, the time savings remain scenario estimates until measured with staff. Likewise, the perfect internal score means only that \claire{} handled the failure types represented in this deterministic synthetic benchmark. It does not establish performance on arbitrary portals, authentication, CAPTCHA, uploads, scanned documents, or asynchronous behavior.

This scope matters because form automation can cause silent operational harm through omitted dependencies, malformed codes, or unsupported values. Such errors may not directly change a diagnosis, but they can delay access, contribute to denial, or create rework. Medical-informatics evaluation should therefore include validation, escalation, traceability, and workflow fit, not field accuracy alone.

\subsection{Workflow integration and governance implications}

A practical deployment would start from an EHR or administrative work item, draft portal or API entries, attach source evidence and validation traces, and return status to the work queue. \claire{} is therefore a staff-review accelerator, not an independent submitter. It must not infer missing facts, alter source meaning, or proceed after unresolved validation; traces should support review, appeals, security, and governance \cite{kelly2019clinicalimpact,lekadir2025futureai,nist2023airmf}. FHIR Questionnaire and QuestionnaireResponse can represent questions, answers, and constraints, while SMART on FHIR supports EHR integration \cite{hl7fhirquestionnaire,hl7fhirquestionnaireresponse,mandel2016smart}. The architecture complements these APIs where portals and exception workflows remain.

\section{Limitations}

This evaluation used deterministic synthetic administrative records and computable schemas. That design protects privacy and makes every expected value reproducible, but it does not capture the full diversity, missingness, ambiguity, coding practices, or distribution shifts of real organizations. Two form families were unchanged controls inside the validation-stress suite, and the represented errors were deliberately chosen to match implemented rules; the perfect full \claire{} score must therefore be interpreted within that constructed coverage. The evaluation also did not test authentication, file uploads, scanned documents, CAPTCHA, production portal latency, or policy changes.

The primary benchmark also does not establish autonomous LLM-agent competence. Although the role modules called a shared language model, the returned model content did not control benchmark actions. The perfect score therefore evaluates the deterministic guarded executor under the represented schemas and perturbations; direct LLM mapping performance is limited to the separate strict-output audit.

The operational time model used assumed task durations rather than observed staff behavior. It cannot establish real productivity, workload, trust, or adoption. The next evidence step is a prospective staff-facing simulation using synthetic or appropriately de-identified cases. Administrative staff, care coordinators, billing specialists, or trained reviewers should compare manual completion with trace-assisted review while measuring observed time, correct submit-or-escalate decisions, unsafe-case blocking, workload, and perceived trust. A later non-production portal study should then test browser failures and workflow integration before any live submission is considered.

\section{Future work}
Future work should progress from public interactive benchmarks to staff-facing tests, non-production portals, and EHR-integrated work queues before live deployment. DECIDE-AI can guide early live evaluation \cite{vasey2022decideai}; CONSORT-AI and SPIRIT-AI would become relevant only for a later trial testing whether \claire{} changes a process of care \cite{liu2020consortai,rivera2020spiritai}.

\section{Conclusion}

\claire{} demonstrates that healthcare administrative form completion can be made more reliable by placing language-model reasoning inside explicit medical-informatics controls. In a 24,000-episode benchmark, schema grounding and validation were necessary for robust performance under interface perturbations and validation-stress conditions. In a 6,000-episode operational simulation, the guarded workflow produced reviewable drafts, focused review, and escalation capture, while calculated time savings of 149.7–165.5 seconds per case remained dependent on explicit timing assumptions. These findings support a design principle for healthcare administrative automation in which language-assisted mapping is embedded within structured information-system controls and remains subject to validation, traceability, human review, and escalation.

\section*{Ethics statement}
This study used synthetic healthcare administrative records and public benchmark-derived answer-key data. No real patient records, protected health information, or human-subject data were used. The experiments did not evaluate diagnosis, treatment recommendation, or patient-facing clinical decision-making. Any deployment on real healthcare data would require institutional privacy, security, compliance, and workflow review.

\section*{CRediT authorship contribution statement}
Garapati Keerthana: Conceptualization, Methodology, Software, Investigation, Data curation, Formal analysis, Writing - original draft, Visualization. Manik Gupta: Supervision, Conceptualization, Methodology, Writing - review and editing.

\section*{Funding}
This research did not receive any specific grant from funding agencies in the public, commercial, or not-for-profit sectors.

\section*{Declaration of competing interest}
The authors declare that they have no known competing financial interests or personal relationships that could have appeared to influence the work reported in this paper.

\section*{Data availability statement}
No patient data or protected health information were used. The synthetic records, computable schemas, and derived benchmark materials supporting the findings are available from the corresponding author on reasonable request.

\section*{Declaration of generative AI and AI-assisted technologies in the manuscript preparation process}
During preparation of this manuscript, the authors used AI-assisted tools for language editing and grammar checks. The authors reviewed and edited the output as needed and take full responsibility for the content of the published article.

\appendix

\section{AI, reproducibility, and implementation transparency appendix}
This appendix provides transparency material for an AI-enabled medical-informatics workflow study, consistent with medical-AI self-assessment guidance \cite{cabitza2021checklist}. \claire{} is a workflow-agent architecture rather than a diagnostic prediction model. The reported system did not train a clinical model on patient data. The 24,000-episode executor called Qwen/Qwen2.5-0.5B-Instruct through each role's prompt interface, but only model-call metadata was logged and returned model content was not used to authorize actions; execution relied on schemas, deterministic matching, validation, and controlled correction. The separate strict-output audit evaluated executable mappings from Qwen2.5-1.5B and Qwen2.5-7B.

\begin{itemize}[leftmargin=*]
  \item \textbf{Task:} Healthcare administrative form completion for intake, prior authorization, benefit verification, credentialing, and claims correction.
  \item \textbf{Input data:} Synthetic structured records and computable form schemas; no patient data or protected health information were used.
  \item \textbf{Models:} The main executor issued prompts to Qwen2.5-0.5B but used rule-governed schema logic for actions; the strict audit used Qwen2.5-1.5B and Qwen2.5-7B for executable source-to-field mapping checks.
  \item \textbf{Evaluation:} Clean and validation-stress suites, four interface variants, strict-output LLM audit, and 6,000-episode operational simulation.
  \item \textbf{Operational metrics:} Staff-reviewable draft rate, focused-review rate, blocked-escalation rate, validator-intercept rate, time saved per case, hours saved, false-reassurance rate, and unsafe-case capture.
  \item \textbf{Human role:} Staff review is required for unresolved ambiguity, missing source support, validation failure, or escalation.
  \item \textbf{Prospective instrument:} A staff-facing pilot protocol and trust survey template were prepared for follow-on evaluation with synthetic or de-identified cases.
  \item \textbf{Deployment status:} Evaluated in reproducible benchmark and operational simulation settings before live clinical or payer deployment.
\end{itemize}

\begin{figure}[!htbp]
\centering
\includegraphics[width=\textwidth]{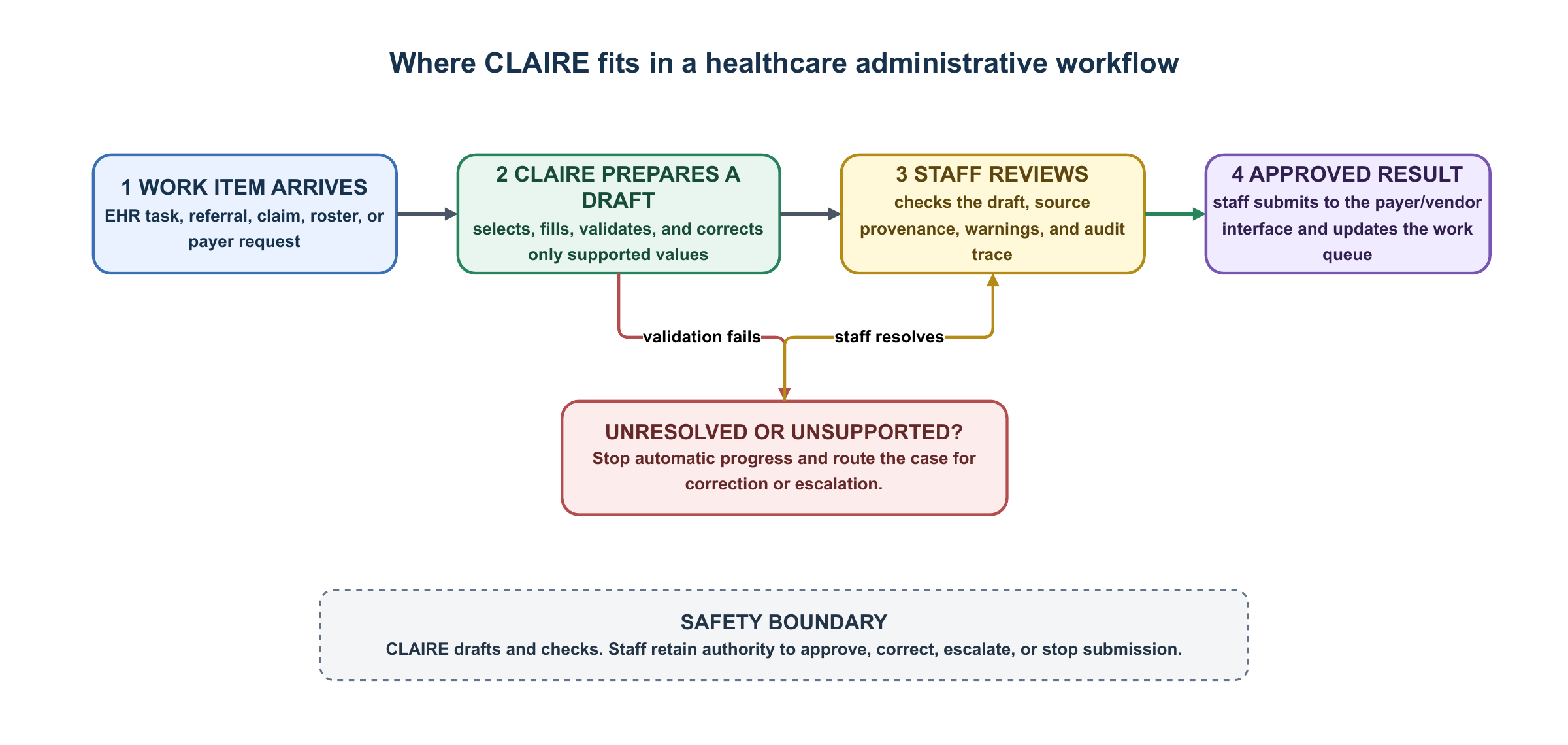}
\caption{Intended workflow placement. \claire{} prepares a validated draft and trace from an incoming administrative work item. Staff retain authority to approve, correct, escalate, or stop submission to a payer or vendor interface.}
\label{fig:workflow}
\end{figure}

\begin{figure}[p]
\centering
\includegraphics[width=0.92\textwidth]{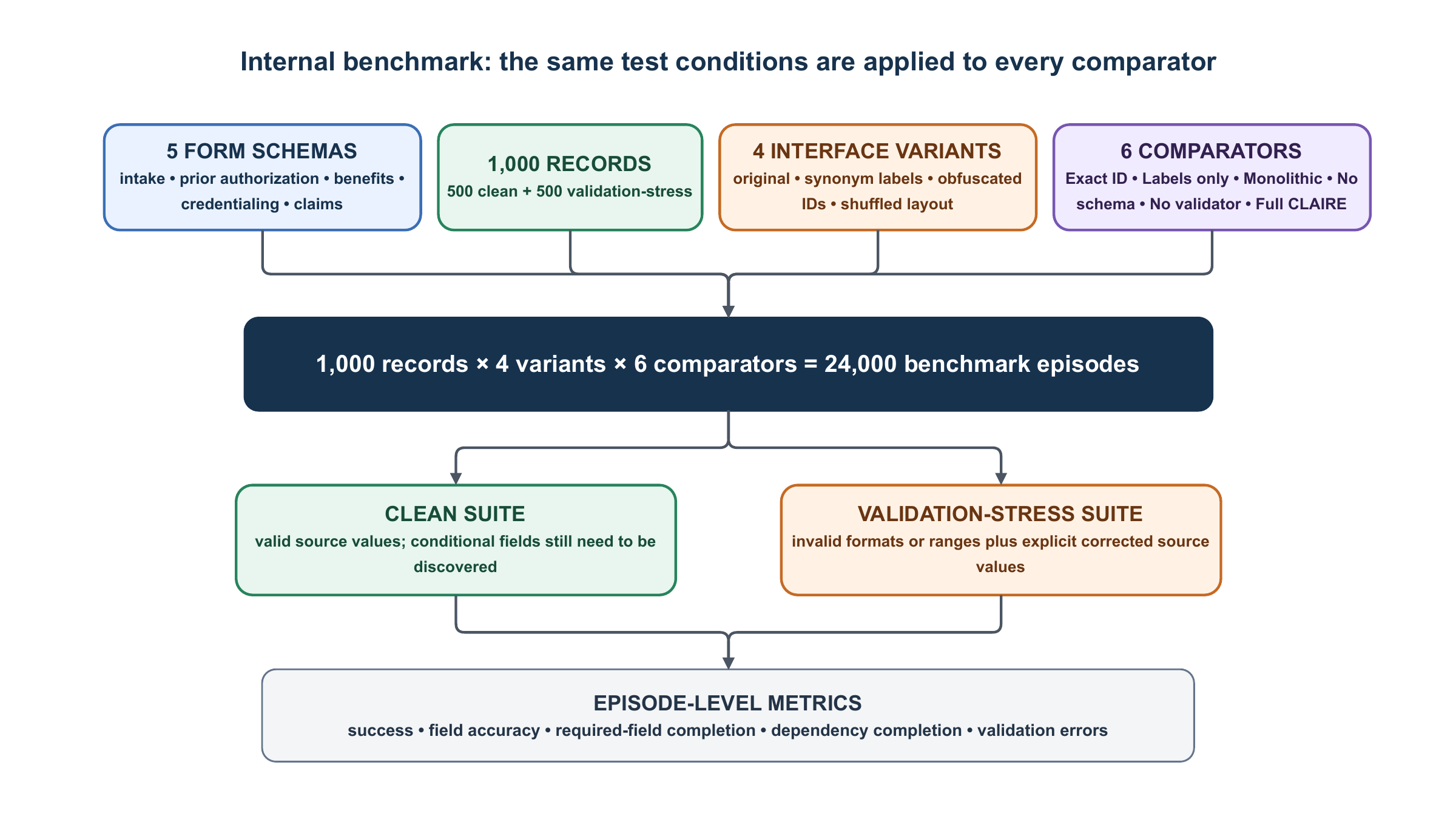}
\caption{Internal benchmark design. The same schemas, records, and interface variants were applied to all six comparators.}
\label{fig:benchmark-data}
\end{figure}

\begin{figure}[p]
\centering
\includegraphics[width=0.92\textwidth]{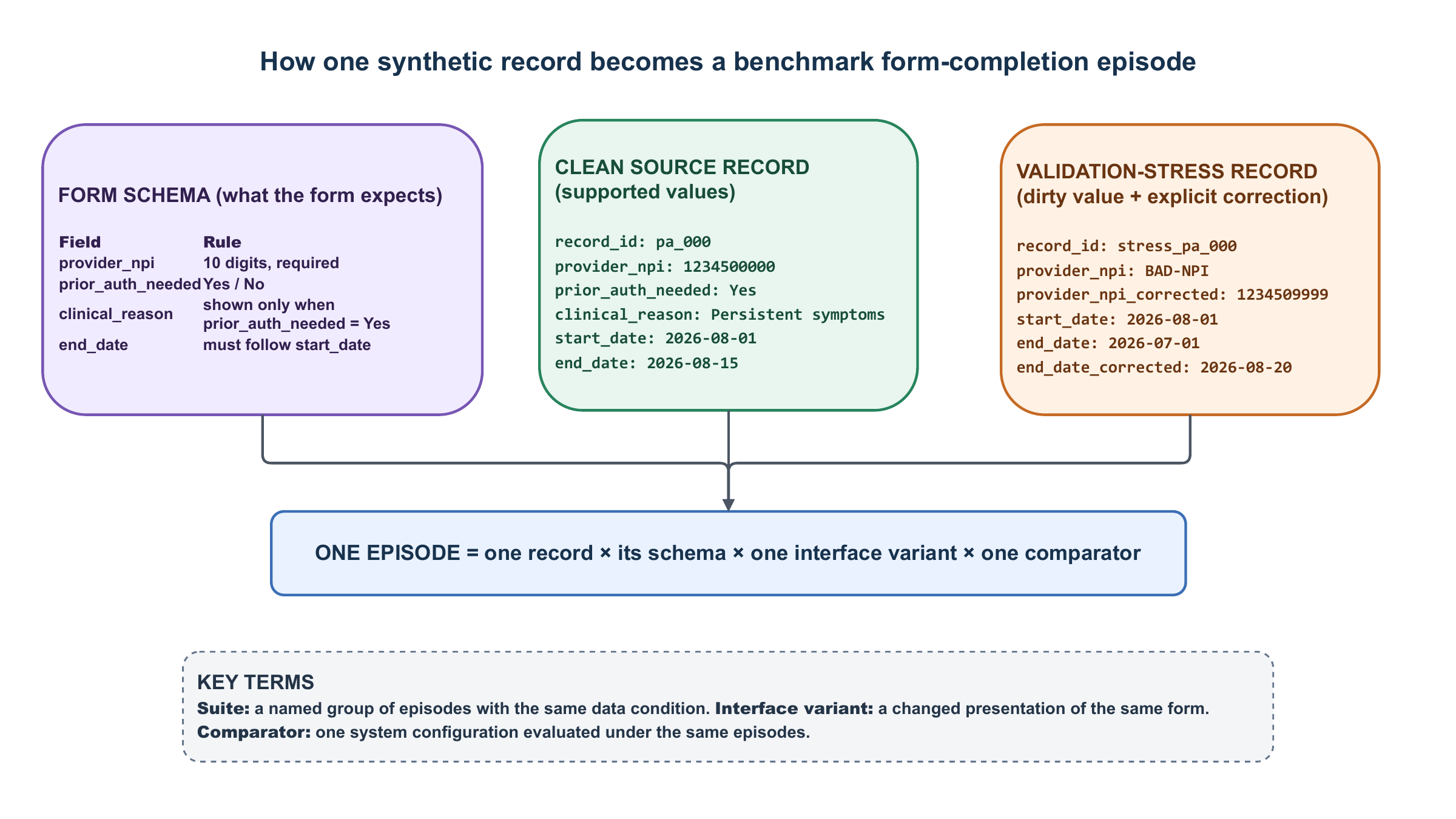}
\caption{Synthetic-data example. The schema defines the target, while the stress record contains invalid values and explicit corrections. All displayed names and identifiers are synthetic.}
\label{fig:dataset-sample}
\end{figure}

\section{External answer-key alignment check}

A secondary check used 230 healthcare-like examples extracted from released FormFactory answer-key data, across four variants and five comparators. Exact ID automation reached 0.750 success. Labels only, Monolithic, \claire{} without schema, and full \claire{} reached 1.000 success because the adapter flattened the answer key into field--value mappings. This result supports semantic field–value alignment but should not be interpreted as a full FormFactory interactive-coordinate reproduction, external clinical validation, or real-world workflow evaluation.

\newpage
\bibliographystyle{plainnat}
\bibliography{references}

\end{document}